# Accelerating Machine Learning via the Weber-Fechner Law

**Balas Natarajan Kausik[1]**

Jan 2022

## Abstract

The Weber-Fechner Law observes that human perception scales as the logarithm of the stimulus. We argue that learning algorithms for human concepts could benefit from the Weber-Fechner Law. Specifically, we impose Weber-Fechner on simple neural networks, with or without convolution, via the logarithmic power series $\sum_{k}(1/k)p_{(k)}^{k}$ of their sorted output $\{p_{(1)}, p_{(2)}, ...\}$. Our experiments show surprising performance and accuracy on the MNIST data set within a few training iterations and limited computational resources, suggesting that Weber-Fechner can accelerate machine learning of human concepts.

[1] Unaffiliated independent: https://www.linkedin.com/in/bnkausik/  Contact: bnkausik@gmail.com

# 1 Introduction

The Weber-Fechner law dates back to the 1800's and empirically observes that human perception and cognition vary as the logarithm of the input stimulus. Basic sensory functions such as vision, weight, and noise level follow this law, e.g., we measure noise level on the logarithmic dB scale. The law also extends to cognitive functions such as distinguishing between numbers, [Moyer and Landauer 1967; Longo and Lourenco, 2007; Mackay, 1963, Staddon, 1978].

Since machine learning is often concerned with learning concepts understood by humans, it is worth asking whether constraining machine learning algorithms to obey Weber-Fechner can accelerate their performance and accuracy. Specifically, we look at the question of whether neural networks can be enhanced by imposing conformance with Weber-Fechner.

In the context of Probably Approximately Correct (PAC) learning, [Valiant, 1984], it is known that weak classifiers can be improved by training on distributions that overweight errors, [Freund and Schapire, 1997]. Here, we present empirical evidence that conformance with the Weber-Fechner law can accelerate the performance and accuracy of simple neural networks that learn concepts understood by humans. Specifically, our experiments show surprising performance and accuracy on the MNIST data set within a few training iterations on an ordinary laptop computer without the aid of special purpose hardware. This suggests that Weber-Fechner can accelerate machine learning of concepts understood by humans.

# 2 Background

Let $B$ be the unit ball in $n$-dimensions, i.e.

$$B = \{x \in R^n, ||x||_2 = 1\}.$$

We consider a function $F: B \to \{0, 1, ..., m\}$ defined as

$$F = argmax(F_i)$$

where $F_i: B \to R$ are the *perception functions* of each class $i$.

Invoking Weber-Fechner on the $F_i$, we have

$$|F_i(x_0) - F_i(x_1)| \simeq - log||x_0 - x_1||$$

Since $||x_0|| = ||x_1|| = 1$,

$$||x_0 - x_1||^2 = 2(1 - x_0 \cdot x_1)$$

It follows that,

$$|F(x_0) - F_i(x_1)| \simeq - log(1 - x_0 \cdot x_1)$$

If we set $p = x_0 \cdot x_1$, we get

$$|F(x_0) - F_i(x_1)| \simeq - log(1 - p) \simeq \sum_{k=1}^{\infty} (1/k)p^k \qquad (1)$$

The series of Equation (1) is known to converge in the open interval $[0, 1)$. Intuitively, the perception functions $F_i$ achieve a maximum at an "ideal" point where $p = 1$. The ideal point may well be a hypothetical abstraction that does not occur in practice.

If $F$ and the perception functions $F_i$ are spatially and/or temporally invariant, it follows that we can replace the dot product in the foregoing with the cross-correlation [convolution] operation[2] and set

$$p = max(x_0 * x_1) = max(x_0 \cdot Tx_1)$$

where the cross-correlation is calculated across the transforms $T$ spanning the spatial and/or temporal dimensions of invariance. Henceforth, we use $a \bullet b$ to denote either $a \cdot b$ or $max(a * b)$.

# 3 Application to Neural Networks

We are given labeled training samples of a multi-class function $F: B \to \{0, 1, ..., m\}$. Our goal is to construct a classifier $C: B \to \{0, 1, ..., m\}$ that minimizes the error $\{C(x) \neq F(x)\}$ on a set of labeled test samples $\{x\}$, independent of the training samples. Our overall multi-class classifier $C$ is below.

$$C(x) = argmax\ (C_i(x))$$

where $C_i: B \to R$ is the perception for class $i$.

Each $C_i$ consists of linear units $\{C_{(1)i}, C_{(2)i}, ... C_{(K)i}\}$ where each unit is specified by a weight vector $v_{(k)i}$ in $B$ such that the output for input $x$ is given either by

$$C_{(k)i} = v_{(k)i} \bullet x$$

with the dot product $v_{(k)i} \cdot x$ or, for spatial or temporal invariance, $max(v_{(k)i} * x)$. There are no hidden units.

For input $x$, we sort the outputs $p_{(*)i} = C_{(*)i} \bullet (x)$ such that

$$p_{(1)i} \geq p_{(2)i} \geq p_{(3)i} \geq ... \geq p_{(K)i}$$

Inspired by Equation (1) above, our perception function for class $i$ is of the form

$$C_i(x) = \sum_{k=1}^{K} \left(\frac{1}{k}\right)\left(p_{(k)i}\right)^k \qquad (2)$$

and the overall classifier is

$$C(x) = argmax(C_i(x)) \qquad (3)$$

We do not know the "ideal" of Equation (1). Hence, Equation (2) uses a composite of the points on the unit ball represented by the units, in the form of Equation (2). Furthermore, a total of 50 to 80 terms in Equation (2) across all classes proved sufficient in our experiments.

---

[2] Please note that the terms cross-correlation and convolution are often used interchangeably in the literature.

# 4 Experiments

We test our algorithm on the MNIST data set of [LeCun et al., 2021]. The data set comprises a training set of 60,000 handwritten digit samples, and an independent test set of 10,000 samples. Each sample is of the form $(x, l)$ where $x$ is a 28x28 pixel map and $l$ is the label indicating the digit represented by the pixel map. Our algorithm constructs a set of linear units $M = \{(v_1, l_1), (v_2, l_2) ...\}$ where each tuple comprises a unit vector $v$ and a label $l$.

---

**Algorithm 1:** Classifier for MNIST Data of $n$ units
**Input:** MNIST Training samples; max number of units $n$
**Output:** Set of linear units $M = \{(v_1, l_1), (v_2, l_2) ..., (v_n, l_n)\}$

initialize $M = \{\}$;
**for** training epochs 1, 2, 3,… **:**
  **for** each training sample $(x, l)$:
    let $l_i =$ output of Algorithm 2 on $x$;
    **if** $l_i \neq l$ :
      **if** *cardinality(M) < n*:
        add $(x, l)$ to $M$;
      **else:**
      **for** $(v, l_i) \in M$ in sorted order of $p_{(k)i}$**:**

$$v = v - (\alpha/k)(v \cdot x) \, x \quad\quad\quad (4)$$

        normalize $v$;

---

In brief, Algorithm 1 greedily adds misclassified training samples as linear units to the set $M$ until the maximum allowed number is reached, and then iterates over the training samples to adjust the weights on the units. At each iteration, the samples are classified by Algorithm 2, which combines the sorted output of the units via Equation (2) to obtain a classification label per Equation (3). Algorithm 1 back-propagates the classification errors via Equation (4), applying a step-size factor $(\alpha/k)$, where $\alpha$ is manually selected.

Our experiments were performed with Numpy & Scipy Python on a Macbook Pro 2.4GhZ/8GB Intel CPU.

---

**Algorithm 2:** Classifier for MNIST Data
**Input:** MNIST test sample $(x, l)$
**Output:** Predicted label $L$ for $x$
**Parameters**: Set of units $M = \{(v_1, l_1), (v_2, l_2) ...\}$

**for** $l \in \{0, 1,... 9\}$:
    let $\{p_{(1)i}, p_{(2)i} ... p_{(K)i}\} = \text{sort}(\{v \bullet x | (v, l) \in M\})$;

$$L = \underset{i}{argmax}\left(\sum_{k=1}^{K}\left(\frac{1}{k}\right)\left(p_{(k)i}\right)^k\right) \quad\quad\quad (5)$$

---

# 5 Experimental Results

In our experiments, we ran Algorithm 1 for 5 training interactions each for

$$n = \{300, 600, 1200, 2400, 4000\}.$$

At *n*=4000, after 3832 units were added, the training error was ~.08% and there were no misclassified training samples that were not already added to the set M. We then ran Algorithm 2 on the test samples, (1) as is (2) convolve 1 pixel in the testing phase only and (3) convolve one pixel and +/- 10 degrees rotation in the testing phase only. The results are shown in Fig. 1 and Table 1.

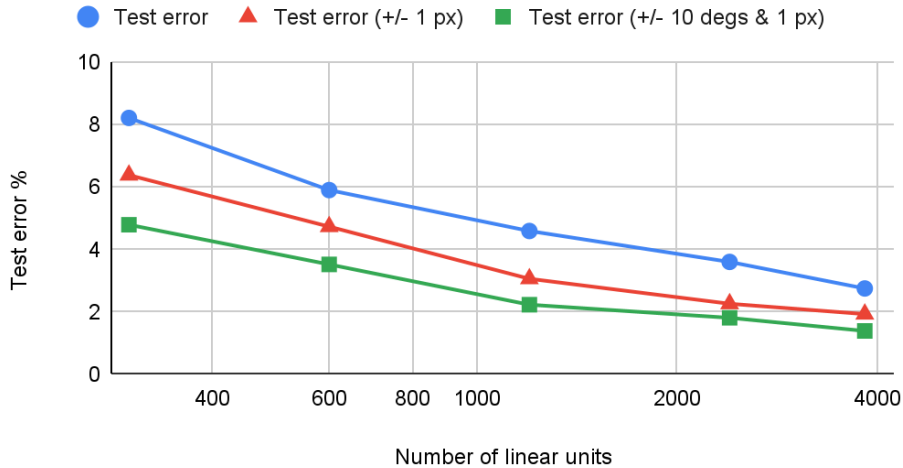

Fig. 1: Test error % after 5 training epochs

We can expand the span of convolution from one pixel shift and +/-10$^0$ rotation to a broader range of shifts and angles, as well as inject elastic distortions into the training samples as proposed in [Ciresan et al. 2012]. Convolving over a broader span of distortions can be carried out during training and/or testing, depending on our relative computational efficiency goals. If convolving during training, Equation (4) of Algorithm 1 becomes

$$v = v - (\alpha/k)(v \cdot Tx)Tx \qquad (6)$$

where *T* ranges over the transforms spanning the space of spatial and/or temporal invariance. We expect expanding the span of convolution can improve the accuracy towards the accuracy reported in Table 2. Results on a range of methods for comparison are documented by [LeCun et al, 2021].
Some work in the literature [Hirata & Takahashi, 2020; An et al, 2020] report the use of ensembles of networks and ensembles of ensembles to improve the accuracy further. [An et al., 2020] report a 95% confidence accuracy of ~0.20%, which is comparable to [Ciresan et al, 2012], but a hand-picked "best accuracy" that is higher.

| # units | Test error | Test Error (+/- 1 px) | Test error (+/-10 degs & 1px) |
|---|---|---|---|
| 300 | 8.21 | 6.38 | 4.79 |
| 600 | 5.9 | 4.73 | 3.52 |
| 1200 | 4.59 | 3.06 | 2.23 |
| 2400 | 3.6 | 2.26 | 1.81 |
| 3832 | 2.75 | 1.93 | 1.39 |

Table 1: Test error % after 5 training epochs

| Network | Preprocessing | Test Error |
|---|---|---|
| committee of 35 conv. net, 1-20-P-40-P-150-10 [elastic distortions] | Width normalization | 0.23 |

Table 2: Test Error % [Ciresan et al, 2012]

Our experiments suggest that enforcing Weber-Fechner on any learning algorithm can accelerate learning of human concepts.

# 6 Summary

The Weber-Fechner Law observes that human perception scales as the logarithm of the stimulus. We present empirical evidence that conformance to Weber-Fechner law(s) can accelerate machine learning of concepts understood by humans.

# 7 Acknowledgments

Thanks to Anoop Bhattacharjya, Forest Baskett, Jay Jawahar and Prasad Tadepalli for their comments and suggestions.